\documentclass[sigconf]{cidr-2026}
\usepackage{graphicx} 
\usepackage{booktabs}
\usepackage{hyperref}
\usepackage{hyperxmp} 
\definecolor{ForestGreen}{RGB}{34,139,34}
\definecolor{Orange}{RGB}{255, 121, 0}
\usepackage{amsmath} 

\newcommand{\todo}[1]{}
\newcommand{\nice}[1]{}
\renewcommand{\todo}[1]{{\color{ForestGreen} TODO: {#1}}}
\makeatletter
\let\new\@undefined
\makeatother
\newcommand{\new}[1]{{\color{black}{#1}}}

\usepackage{xcolor}
\usepackage{dblfloatfix}
\usepackage{diagbox}
\usepackage{soul}
\usepackage{tikz}

\begin{document}

\title{BenchPress: A Human-in-the-Loop Annotation System \\
for Rapid Text-to-SQL Benchmark Curation} 

\author{
Fabian Wenz$^{1,2}$\quad
Omar Bouattour$^{1,2}$\quad
Devin Yang$^{2}$\quad
Justin Choi$^{2}$\quad
Cecil Gregg$^{2}$\\
Nesime Tatbul$^{3,2}$\quad
Çağatay Demiralp$^{4,2}$
}
\affiliation{
$^{1}$TU Munich\quad
$^{2}$MIT\quad
$^{3}$Intel Labs\quad
$^{4}$AWS AI Labs
}
\email{
{fab_wenz, kevin023, jchoi, cgregg4}@mit.edu,
omar.bouattour@tum.de,
{tatbul, cagatay}@csail.mit.edu
}

\renewcommand{\shortauthors}{Wenz et al.}
\newcommand{\fabian}[1]{\textcolor{orange}{\textbf{Fabian:} #1}}


\begin{abstract}
Large language models (LLMs) have been successfully applied to many tasks, including text-to-SQL generation. However, much of this work has focused on publicly available datasets, such as \textsc{Fiben}, \textsc{Spider}, and \textsc{Bird}. Our earlier work showed that LLMs are much less effective in querying large private enterprise data warehouses, releasing \textsc{Beaver}, the first private enterprise text-to-SQL benchmark. To create \textsc{Beaver}, we leveraged SQL logs, which are often readily available. However, manually annotating these logs to identify which natural language questions they answer is a daunting task. Asking database administrators, who are highly trained experts, to take on additional work to construct and validate corresponding natural language utterances is not only challenging but also quite costly. 

To address this challenge, we introduce \textbf{BenchPress}, a human-in-the-loop system designed to accelerate the creation of domain-specific text-to-SQL benchmarks. Given a SQL query, BenchPress uses retrieval-augmented generation (RAG) and LLMs to propose multiple natural language descriptions. Human experts then select, rank, or edit these drafts to ensure accuracy and domain alignment. We evaluated BenchPress on annotated enterprise SQL logs, demonstrating that LLM-assisted annotation drastically reduces the time and effort required to create high-quality benchmarks. Our results show that combining human verification with LLM-generated suggestions enhances annotation accuracy, benchmark reliability, and model evaluation robustness. By streamlining the creation of custom benchmarks, BenchPress offers researchers and others a mechanism for assessing text-to-SQL models on a given domain-specific workload\new{, where no comparable tools exist today}. BenchPress is freely available via our public GitHub repository\footnote{\url{https://github.com/fabian-wenz/enterprise-txt2sql}} and accessible for use on our website\footnote{\url{http://dsg-mcgraw.csail.mit.edu:5000/}}.


\end{abstract}


\maketitle

\section{Introduction}\label{sec:introduction}
The adoption of large language models (LLMs) for text-to-SQL conversion has gained traction in enterprise settings, where databases are vast, but expert annotation resources are scarce. While academic research has produced powerful text-to-SQL models, enterprises face a critical problem: \textit{how well do these models perform on their data?} 

Public benchmarks like \textsc{Spider}~\cite{yu2018spider}, \textsc{Bird}\cite{wang2023bird}, and \textsc{Fiben} \cite{sen2020athenaplusplus} provide valuable testbeds for general-purpose text-to-SQL evaluation, but they fail to capture enterprise-specific challenges, such as:
\begin{itemize}
    \item \textbf{Schema complexity and ambiguity}: Enterprise databases are often heterogeneous, integrating tables from different systems with overlapping but inconsistent naming conventions.
    \item \textbf{Domain-specific terminology}: Public datasets lack the specialized vocabulary and abbreviations used in finance, healthcare, IT, and other industries.
    \item \textbf{Privacy and security constraints}: Unlike academic benchmarks, enterprise SQL logs cannot be publicly shared, making it difficult for organizations to benchmark models against real-world queries.
\end{itemize}

\definecolor{upgreen}{RGB}{0,150,0}
\definecolor{downred}{RGB}{180,0,0}
\newcommand{\up}[1]{\textcolor{upgreen}{$\uparrow$\,#1}}
\newcommand{\upbutdown}[1]{\textcolor{downred}{$\uparrow$\,#1}}
\newcommand{\down}[1]{\textcolor{downred}{$\downarrow$\,#1}}

\begin{figure}[ht]
    \centering
    \includegraphics[width=\linewidth]{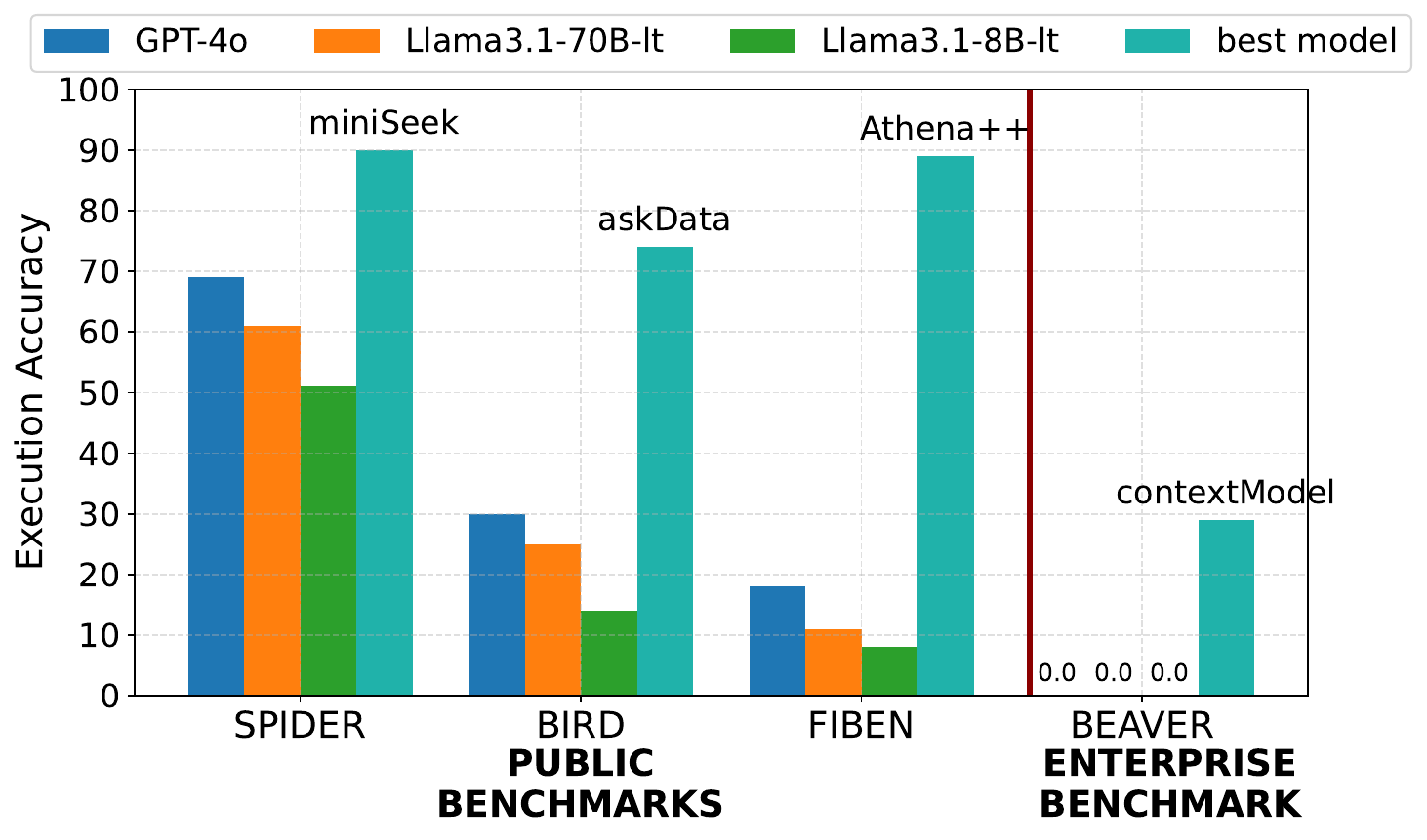}
    \caption{Execution accuracy of different LLMs on public benchmarks (\textsc{Spider}, \textsc{Bird}, \textsc{Fiben}) and the enterprise benchmark (\textsc{Beaver}). Since the best-performing model varies across datasets, the specific model achieving the highest accuracy is labeled above each \textcolor{teal}{teal}-colored bar: \textsc{Spider} – miniSeek~\cite{miniseek2023}, \textsc{Bird} – askData~\cite{askData2025}, \textsc{Fiben} – Athena++ \cite{sen2020athenaplusplus}, and \textsc{Beaver}\cite{chen2024beaver} – contextModel.
}
    \label{fig:exec-accuracy}
\end{figure}

As a result, companies risk deploying models that fail on their data due to domain mismatch, leading to unreliable query generation and poor automation performance. While recent LLMs such as GPT-4o and fine-tuned LLaMA variants achieve impressive results on public datasets like \textsc{Fiben}, \textsc{Spider}, and \textsc{Bird}—to the point that the text-to-SQL task may appear nearly solved—, their execution accuracy \footnote{Execution accuracy measures whether the result of executing the predicted SQL query matches that of the gold SQL \cite{yu2018spider, wang2023bird}.} drops sharply on enterprise datasets such as \textsc{Beaver}\cite{chen2024beaver}. Figure~\ref{fig:exec-accuracy} visualizes this gap, showing a dramatic execution accuracy drop when the same models are evaluated on real-world enterprise queries. This discrepancy highlights the limitations of existing public benchmarks and the risk of overestimating model readiness for production use. To avoid deployment failures, enterprises must evaluate model performance under their own schemas, domain-specific terminology, and query patterns.

To address this gap, we introduce \textbf{BenchPress}, a novel system designed to enable organizations to create their own text-to-SQL benchmarks quickly and efficiently. \new{Unlike prior work, BenchPress focuses on the SQL-to-NL annotation problem itself---a capability largely absent in existing tools and particularly important in enterprise settings, where SQL logs are abundant but corresponding natural language descriptions are scarce.} 
Given a SQL query, BenchPress generates a natural language (NL) description, allowing domain experts to review and refine it, significantly reducing annotation effort. 
It offers a scalable and privacy-aware foundation for enterprise-grade text-to-SQL evaluation.

By integrating LLM-generated suggestions with human validation, BenchPress accelerates annotation while maintaining accuracy, allowing enterprises to benchmark any text-to-SQL model against their own private datasets. We evaluated BenchPress on enterprise SQL logs, demonstrating its effectiveness in generating high-quality, domain-specific training data while reducing manual effort. 

Beyond enabling benchmark creation, BenchPress has broader implications for enterprise AI adoption, allowing organizations to:
\begin{itemize}
    \item \textbf{Assess model generalization} to proprietary schemas before deployment.
    \item \textbf{Optimize fine-tuning strategies} by identifying failure cases.
    \item \textbf{Improve interpretability} of text-to-SQL models by systematically evaluating outputs.
\end{itemize}

By providing a scalable, adaptable, and privacy-aware solution for enterprise text-to-SQL benchmarking, BenchPress paves the way for more robust and domain-specific LLM evaluations.

\section{Related Work}

In this section, we summarize related work in benchmarks, annotation tools, LLM-based SQL generation, and enterprise adaptation.

\paragraph{Text-to-SQL Benchmarks}  
Public benchmarks such as \textsc{Spider} ~\cite{yu2018spider}, \textsc{Bird} \cite{wang2023bird}, and \textsc{Fiben} \cite{sen2020athenaplusplus} have driven progress in general-purpose text-to-SQL tasks. While these datasets capture diverse query types, they focus on clean, publicly available schemas. The \textsc{Beaver} benchmark \cite{chen2024beaver} introduces a combined corpus from academic and enterprise-inspired sources, including the network datasets and the datawarehouse from the Massachusetts Institute of Technology. \textsc{Beaver} demonstrates that LLMs struggle with more complex queries and heterogeneous schemas, motivating the need for domain-specific evaluation and adaptation tools. Unlike \textsc{Beaver}, which centers on model benchmarking, BenchPress focuses on enabling rapid benchmark creation through LLM-assisted human annotation.

\paragraph{LLMs for SQL Generation}  
Large language models such as Codex, GPT-4, and DeepSeek have shown strong results in translating natural language into SQL \cite{chen2021evaluating, rajkumar2022evaluating}. However, their performance degrades significantly in domain-specific or enterprise contexts. Maamari et al.~\cite{maamari2025cidr} highlight this gap by studying LLMs in enterprise scenarios, finding that pre-trained models often fail due to domain-specific vocabulary and schema ambiguity. While their focus is on improving LLM robustness through better model training and prompting, our work complements this by tackling the data creation bottleneck—providing a system that enables enterprises to efficiently construct accurate, workload-specific training and evaluation data.

\paragraph{Annotation Systems and Tools}  
Several systems have explored semi-automated dataset creation. Andrejczuk et al.~\cite{wang2022sato} propose schema-aware data-to-text generation pipelines, and Xu et al.~\cite{yao2023sql2text} study SQL-to-text generation using encoder-decoder models. These approaches are primarily model-driven and assume public or synthetic data. In contrast, BenchPress supports human-in-the-loop workflows designed specifically for private enterprise logs, integrating LLM-generated suggestions with domain expert review.

\paragraph{Enterprise Data Challenges}  
Enterprise databases differ from academic benchmarks in terms of scale, schema complexity, and sensitivity. Prior work on federated training and schema linking \cite{shi2022learning, zhong2020semantic} has explored solutions for isolated aspects, but few systems address the full annotation pipeline. BenchPress fills this gap by enabling benchmark creation that is both domain-adaptive and privacy-aware—without requiring public data release or costly in-house labeling from scratch.

\paragraph{BenchPress in Context}  
Unlike prior systems, BenchPress provides a practical toolkit for constructing domain-specific text-to-SQL benchmarks. \new{The current version primarily focuses on SQL-to-NL annotation, which serves multiple downstream purposes including validation, semantic enrichment, and human verification; thereby enabling scalable and robust benchmarking in enterprise settings.} BenchPress provides a means to generate new, customized corpora tailored to real-world workloads.

\section{Enterprise SQL Logs: Challenges}\label{sec:Enterprise SQL Logs: Challenges}



\begin{table*}[h!]
\centering
\begin{tabular}{lcccccc}
\toprule
\textbf{Query Sets} & \textbf{\#Keywords} & \textbf{\#Tokens} & \textbf{\#Tables} & \textbf{\#Columns} & \textbf{\#Agg} & \textbf{\#Nestings} \\
\midrule
\textbf{BEAVER (DW)} & 15.6 & 99.8 & 4.2 & 11.9 & 5.5 & 2.05 \\
\midrule
Spider & \down{80.8\%} & \down{81.5\%} & \down{64.3\%} & \down{75.6\%} & \down{83.6\%} & \down{45.5\%} \\
FIBEN & \down{39.1\%} & \up{62.2\%} & \down{9.5\%} & \down{18.5\%} & \down{63.6\%} & \down{23.8\%} \\
BIRD & \down{73.1\%} & \down{68.7\%} & \down{54.7\%} & \down{63.0\%} & \down{87.3\%} & \down{45.5\%} \\
\bottomrule
\end{tabular}
\caption{\new{Query-level complexity metrics across benchmarks: \\
This table compares structural and linguistic query characteristics between BEAVER's Datawarehouse (DW) and existing benchmarks (Spider, FIBEN, BIRD). Percentages indicate the relative difference compared to BEAVER (DW).}}
\label{tab:query-complexity}
\begin{tabular}{lcccccc}
\toprule
\textbf{Data Sets} & \textbf{Columns/Table} & \textbf{Rows/Table} & \textbf{Table/DB} & \textbf{Uniqueness} & \textbf{Sparsity} & \textbf{Data Types} \\
\midrule
\textbf{BEAVER (DW)} & 15.6 & 128K & 99 & 45.9\% & 15.0\% & 4 \\
\midrule
Spider & \down{65.4\%} & \down{98.4\%} & \down{94.8\%} & \upbutdown{59.5\%} & \down{100\%} & 0.0\% \\
FIBEN & \down{84.0\%} & \down{40.6\%} & \up{53.5\%} & \upbutdown{28.1\%} & \down{100\%} & \upbutdown{100\%} \\
BIRD & \down{56.4\%} & \up{328.9\%} & \down{54.7\%} & \upbutdown{72.8\%} & \down{100\%} & \upbutdown{74\%} \\
\bottomrule
\end{tabular}
\caption{\new{Data-level complexity metrics across benchmarks: \\
This table reports schema and data characteristics comparing BEAVER's DW to Spider, FIBEN, and BIRD. Percentages show the relative difference compared to the DW section of BEAVER.}}
\label{tab:data-complexity}
\end{table*}

In this research, we have worked with four text-to-SQL benchmarks: \textsc{Spider}, \textsc{Bird}, and \textsc{Fiben} as public datasets, and \textsc{Beaver} as a private, enterprise-oriented dataset. \textsc{Beaver} is based on SQL logs from enterprise databases across industries such as education, technology, and manufacturing. These logs span in total over 300 schemas and nearly 4000 queries, featuring complex, multi-source schemas with inconsistent naming conventions and semantic overlaps. For example, in the MIT data warehouse, a single natural language query can often be answered by multiple SQL queries due to the presence of materialized views and semantic ambiguity across tables. \new{This is visualized in Tables~\ref{tab:query-complexity} and~\ref{tab:data-complexity}.}

\new{In Tables~\ref{tab:query-complexity} and \ref{tab:data-complexity}, the arrows denote the \emph{relative change} of each benchmark with respect to the \textsc{Beaver} (DW) workload: a downward arrow (\down{}) indicates a decrease and an upward arrow (\up{}) indicates an increase relative to \textsc{Beaver} (DW). The color coding highlights whether this change represents an \emph{easier} (\textcolor{downred}{red}) or \emph{harder} (\textcolor{upgreen}{green}) characteristic for the text-to-SQL task. Importantly, higher values are not universally ``better'' or ``worse''; their desirability depends on how they influence model difficulty.}

Next, we discuss the challenges of working with these enterprise SQL logs in contrast to those from open-domain benchmarks:

\paragraph{Domain-Specific Terminology:}
Enterprise SQL logs often contain specialized vocabulary that requires deep contextual understanding. For instance, terms like “J-term” (a one-month January term) are specific to the MIT academic calendar and may be incomprehensible to annotators or models without MIT-specific knowledge. Without domain expertise, LLMs frequently fail to map such terms to the correct database fields, reducing annotation and generation accuracy.

\paragraph{Query Complexity:}
Queries in enterprise settings are substantially more complex than those found in public benchmarks. They commonly include nested subqueries, aggregation dependencies, and recursive joins. A single enterprise query may aggregate data from 5–10 tables and use the result in a nested filter. LLMs, which are typically trained on simpler academic benchmarks, struggle to handle this structural depth.  
\new{In Table~\ref{tab:query-complexity}, higher counts of keywords, nestings, aggregations, and referenced tables all quantify this complexity. Each of these dimensions reflects a form of compositional depth that increases the cognitive load for annotators and the reasoning burden on LLMs, contributing to the substantial enterprise gap observed across workloads.}

\paragraph{Privacy Constraints:}
Enterprise data often contains sensitive or proprietary information, such as employee salaries or internal performance metrics. Because of strict privacy constraints, this data cannot be and traditionally has not been publicly released, which in turn makes it challenging to build models, create representative benchmarks, and evaluate performance accurately. Consequently, organizations must evaluate any model in a secure, private environment using their own in-domain data before deployment.



\paragraph{Schema Ambiguity and Duplication:}
Enterprise data warehouses often aggregate tables from various internal systems, leading to schema inconsistencies. It is common to find multiple tables with identically named columns such as “user\_id” that actually refer to different entities. Disambiguating such cases requires either careful schema documentation or intelligent annotation, both of which pose challenges for LLM-based systems.
\new{In Table~\ref{tab:data-complexity}, lower schema \emph{uniqueness}, i.e., more repeated or semantically similar column names across tables, directly reflects this ambiguity. Because models must distinguish between near-identical attributes, a decrease in uniqueness (\textcolor{upgreen}{$\downarrow$}) is marked as \textcolor{upgreen}{green}, indicating increased benchmark difficulty. 
Similarly, lower \emph{data-type diversity} amplifies ambiguity by reducing the semantic cues available for distinguishing column roles; when tables use only few data types (e.g., diversity near 0.0), LLMs lose an important signal for disambiguation.}

\begin{figure*}[!b]
\centering
\includegraphics[width=\linewidth]{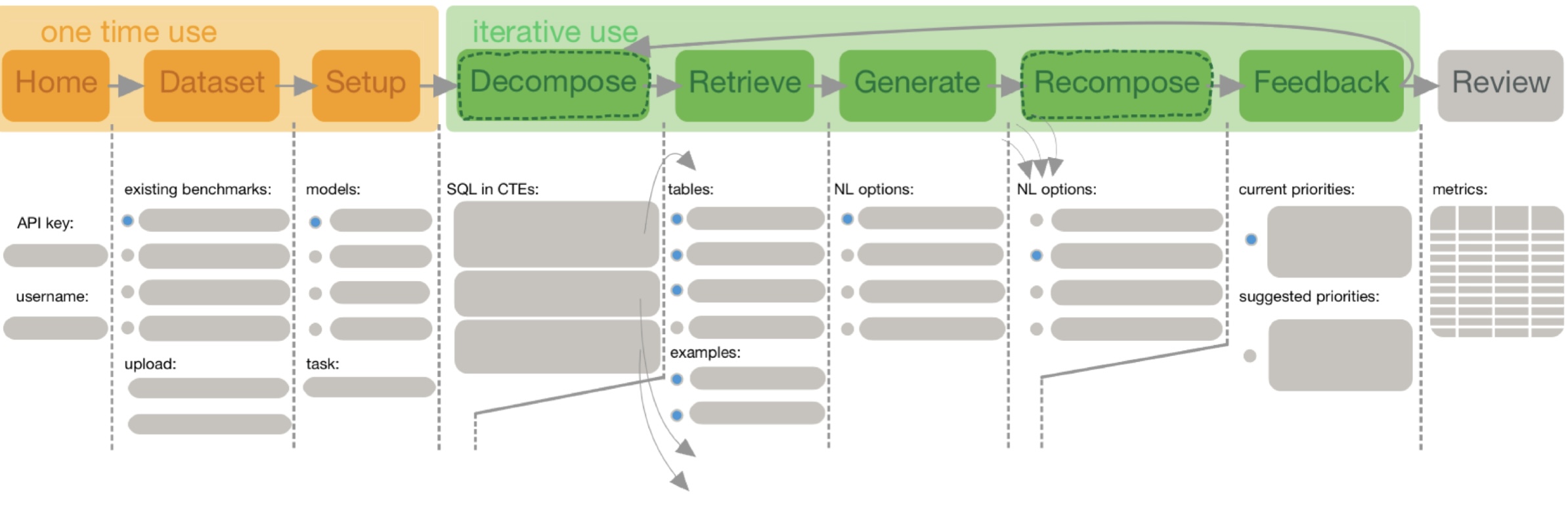}
\caption{BenchPress workflow: initial one-time setup (orange), iterative annotation loop (green), and final review (gray).}
\label{fig:benchpress-flow}
\end{figure*}

\paragraph{Data Sparsity and Imbalance:}
Many enterprise datasets suffer from sparsity and imbalance. 
\new{We observe this not only in \textsc{Beaver}, but also in a large-scale Intel-internal data warehouse, in which real-world operational data exhibits uneven field completeness, heterogeneous logging practices, and highly skewed attribute distributions. For example, the Intel data warehouse\cite{abs-2503-09957} includes performance metrics from millions of devices, yet many fields may be partially or entirely missing. LLMs trained on uniformly structured or synthetic data often struggle with these irregularities, resulting in biased performance on common patterns and failures on rare or incomplete cases.}
\new{In Table~\ref{tab:data-complexity}, these effects are captured by higher schema \emph{sparsity}, meaning a greater proportion of NULL or missing values per column. Additionally, a higher number of columns per table and a larger number of rows per table indicate greater structural and statistical imbalance: models must process wider and deeper tables, often with uneven population across attributes. All of these factors increase annotation and modeling difficulty and are therefore marked as preferable (\up{}) in \textcolor{upgreen}{green} for benchmarking purposes.}

In summary, enterprise SQL logs differ fundamentally from public benchmarks. The challenges span domain-specific terminology, complex query structures, privacy restrictions, ambiguous schemas, and sparse or imbalanced data. Addressing these issues demands tools like BenchPress that incorporate human-in-the-loop annotation, secure evaluation workflows, and domain-aware disambiguation mechanisms.

\section{The BenchPress System}

BenchPress is a human-in-the-loop system designed to accelerate the annotation of SQL logs for building high-quality text-to-SQL benchmarks. It enables domain experts to generate natural language descriptions for SQL queries more efficiently by combining retrieval-augmented generation (RAG), prompt-based LLM outputs, and iterative feedback refinement in a modular and interactive interface.

\subsection{Workflow Overview}

Figure~\ref{fig:benchpress-flow} presents the high-level workflow of BenchPress, which consists of a one-time setup phase followed by a repeated annotation loop. For non-nested SQL queries, the pipeline follows the standard sequence of steps, excluding the dashed steps. For nested queries, BenchPress automatically inserts two intermediate steps—decomposition and recomposition—to improve annotation accuracy and reduce complexity. These optional steps are depicted in the only dashed boxes in the diagram.

\textit{One-Time Setup:}

\begin{enumerate}
    \item \textbf{Project Setup:} \new{The API key is kept entirely within the user's browser storage, ensuring that it never leaves the local client. 
The username serves as a local workspace identifier under which annotation projects are organized. 
Users can maintain multiple projects, each associated with a specific schema and the SQL logs or queries they upload.}
\new{\item \textbf{\new{Dataset Ingestion:}} 
Users upload SQL logs and schema files or select from one of four supported benchmarks:
\textsc{Bird}, \textsc{Fiben}, \textsc{Spider}, or \textsc{Beaver}.
\new{Because SQL logs and schemas can be large, they are stored on the server rather than in the browser.
Standard browser storage mechanisms such as \texttt{localStorage} typically allow only about 5--10\,MB per origin 
~\cite{mdn-webstorage}, and although IndexedDB can support larger datasets, the quota is highly variable across browsers and cannot guarantee availability for large enterprise SQL logs 
~\cite{mdn-indexeddb-quota}. 
Centralized server-side storage ensures the full dataset is always available for retrieval-augmented generation (RAG), which requires global access to all uploaded documents for vector search and efficient retrieval.}
The system then parses and stores this data for further processing.}

    \item \textbf{Task Configuration:} Users choose the annotation direction \new{task} (currently only SQL-to-NL), as well as the language model (e.g., GPT-4o, GPT-3.5 Turbo, or DeepSeek).
\end{enumerate}

\begin{figure*}[t]
    \centering
    \includegraphics[trim={0 3.5cm 0 5cm}, clip, width=\linewidth]{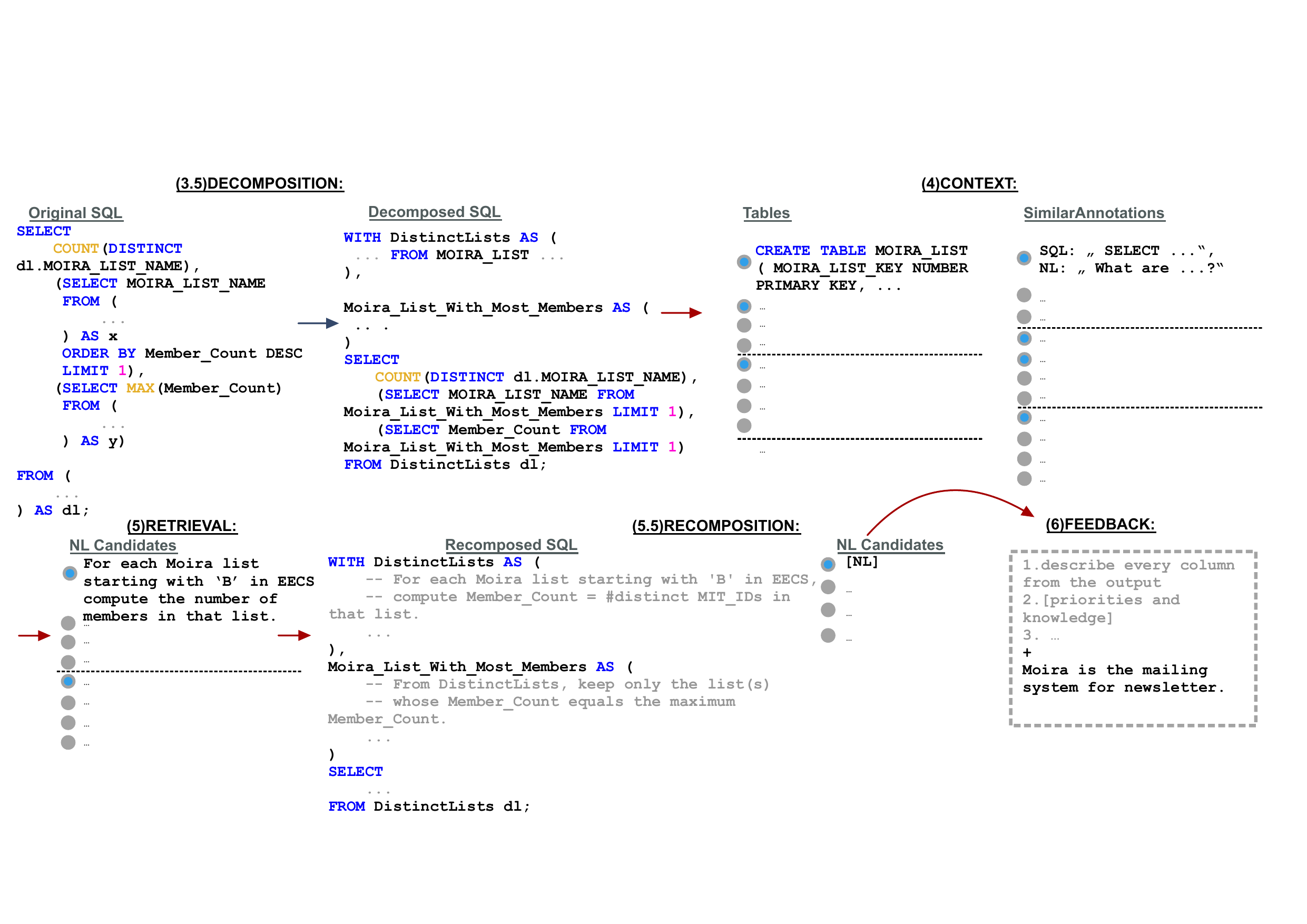 }
    \caption{\textbf{Example annotation run in BenchPress.}
    This figure illustrates how a single SQL query moves through the annotation loop.
    For nested queries, BenchPress performs optional \emph{decomposition} into CTEs, followed by schema and example retrieval, LLM-based NL candidate generation, optional \emph{recomposition} into a unified explanation, and human \emph{feedback}. 
    This example corresponds directly to steps 3.5–6 in the Annotation Loop.}
    \label{fig:benchpress-example}
\end{figure*}

\textit{Annotation Loop:}
To illustrate how a single query progresses through the annotation loop, Figure~\ref{fig:benchpress-example} shows a concrete example passing through decomposition, retrieval, candidate generation, recomposition, and feedback.

\begin{enumerate}
    \setcounter{enumi}{3}
    \item[(3.5)] \textbf{(Optional) Decomposition:} For nested SQL queries, the system rewrites the query into a series of Common Table Expressions (CTEs), breaking it down into \new{semantically logical} subqueries.
    
    \item \textbf{Context Retrieval:} For each SQL query or subquery, the system retrieves semantically similar examples using dense vector embeddings (e.g., Sentence-BERT~\cite{reimers2019sentence}). These examples consist of prior annotated queries (which naturally grow over time) and serve as guidance for generating relevant phrasing. In addition, BenchPress retrieves relevant tables with all their columns from the schema—either via SQL parsing (e.g., using sqlglot) or using the same embedding-based retrieval mechanism as for the examples. This combined context grounds the model's output in both content and structure~\cite{lewis2020retrieval}.
    
    \item \textbf{Candidate Generation:} A large language model (LLM), selected in step 3, generates four candidate natural language descriptions for each SQL query or subquery. The prompt incorporates the retrieved examples and schema context from step 4 using a retrieval-augmented, few-shot prompting approach. Since examples can be long and dilute the prompt, the system always includes the relevant tables but only suggests the top-k retrieved examples to the user. This setup balances informativeness with prompt efficiency. We chose four candidates to balance linguistic diversity with annotation efficiency. Prior work in instruction tuning and human preference modeling often adopts this number as it provides sufficient variation while keeping the cognitive load manageable for human reviewers~\cite{ouyang2022training, wang2022self}. Generating multiple outputs also supports downstream use cases such as ranking, majority voting, or active learning.
    
    \item[(5.5)] \textbf{(Optional) Recomposition:} If decomposition was performed, the system automatically merges the subquery-level descriptions into a single coherent explanation of the original nested SQL query.
    
    \item \textbf{Feedback:} \new{Annotators can rank, refine, discard, or add priorities assigned to the LLM, and can also introduce external domain knowledge or highlight common failure patterns that the model typically misses. This human-in-the-loop feedback loop not only corrects the LLM’s current reasoning but also enriches it with enterprise-specific insights, thereby improving prompt quality over time and providing strong supervision for future fine-tuning~\cite{christiano2017deep}.}
    
    \item \textbf{Review and Export:} If ground truth annotations exist, outputs can be evaluated using automatic metrics (e.g., exact match, BLEU). Otherwise, users rely on qualitative assessment. Final annotations are exported in benchmark-ready format for training or evaluation.
\end{enumerate}
The final exported annotations are then available in the typical benchmark format, i.e., JSON format, for downstream training and evaluation.

\subsection{Key Design Features}

In its design, BenchPress brings together several core techniques and strategies to support accurate and efficient annotation at scale:

\begin{itemize}
    \item \textbf{Retrieval-Augmented Generation (RAG):} To improve relevance, BenchPress retrieves semantically similar SQL queries and their annotations using dense vector search (e.g., via Sentence-BERT~\cite{reimers2019sentence}). These examples are embedded into prompts to ground the model in realistic phrasing patterns and schema usage~\cite{lewis2020retrieval}. This approach aligns with best practices from the seminal Retrieval-Augmented Generation work~\cite{lewis2020retrieval}, which shows significant accuracy gains for specialized NLP tasks through retrieval-enhanced prompting.

    \item \textbf{Prompt Engineering and Refinement:} BenchPress utilizes structured prompt templates tailored explicitly to enterprise SQL logs. When initial suggestions fail to capture the intended meaning, annotators can iteratively refine prompts (e.g., emphasizing "filtering logic") to guide re-generation. This feedback-driven refinement loop has been shown to improve prompt quality and annotation accuracy, mirroring strategies from reinforcement learning from human feedback (RLHF) and human preference modeling~\cite{christiano2017deep}.

    \item \textbf{Query Decomposition:} For nested or structurally complex SQL queries, BenchPress decomposes them into simpler sub-queries. Natural language descriptions are then generated independently and later reassembled, reducing cognitive load and improving annotation precision.

    \item \textbf{Human-in-the-loop Feedback:} \new{At its core, BenchPress emphasizes continuous human oversight. Domain experts iteratively review, edit, rank, discard, or add priorities to the annotations proposed by the LLM, and can inject external domain knowledge or highlight common failure modes that the model typically misses. This structured, iterative feedback loop ensures that annotations meet enterprise-quality standards and prevents error propagation downstream. The approach aligns with Google’s PAIR principles for responsible AI~\cite{dixon2023pair} and is consistent with findings that combining human judgment with AI outputs leads to higher quality, robustness, and trustworthiness~\cite{amershi2019guidelines}.}
\end{itemize}

This modular architecture enables BenchPress to support diverse workflows and database schemas with minimal reconfiguration, while explicitly leveraging human expertise to maximize accuracy and adaptability for enterprise use cases.

\section{User Study}
\subsection{Setup}
To evaluate the impact of BenchPress on annotation efficiency and quality, we conducted a controlled user study using a between-subjects experimental design\new{, meaning that each participant experiences only one of the available conditions. This design prevents learning effects, familiarity with the queries, or fatigue from carrying over between interfaces.} In this design, each participant is randomly assigned to one condition only\new{, ensuring that observed performance differences reflect the properties of the interface rather than improvements from repeated exposure}.
This setup is widely used in HCI and behavioral research, including in experimental frameworks~\cite{HCI2017}.

\new{Our study includes 18 participants, which is well within the range commonly accepted in HCI and empirical systems research. Prior work in usability testing has repeatedly shown that relatively small samples (5--20 participants) are sufficient to uncover the vast majority of issues and yield stable performance differences between interfaces. Nielsen's foundational work demonstrates that diminishing returns set in rapidly after 15 participants~\cite{nielsen1994usability}, while Virzi provides empirical evidence that additional users contribute only marginal increases in explanatory power~\cite{virzi1992refining}. Faulkner further shows that studies with 5--20 users achieve stable, reliable results across usability metrics~\cite{faulkner2003beyond}. Similarly, the ``small-N, high-signal'' evaluation paradigm used throughout systems-HCI research (e.g., in CHI, UIST, and CSCW) relies on focused studies with 12--20 participants to draw statistically meaningful conclusions about task performance, efficiency, and error rates. 

These 18 participants were then }
first grouped into two strata -- \textit{advanced} and \textit{non-advanced} SQL users-based on a pre-study questionnaire assessing their experience and familiarity with relational databases. Within each stratum, participants were randomly assigned to one of three experimental conditions using a balanced Latin square design to ensure counterbalancing:

\begin{itemize}
    \item \textbf{Group A (BenchPress)}: Used the BenchPress interface, including schema information, example tables, logs, and four LLM-generated natural language suggestions per SQL query.
    \item \textbf{Group B (Manual)}: Provided only with schema files and logs, no LLM or suggestion support.
    \item \textbf{Group C (Vanilla LLM)}: Allowed to use a general-purpose LLM (e.g., ChatGPT) via its standard UI, but without RAG-based support or task-specific integration.
\end{itemize}

\new{Each participant was assigned the same set of 30 SQL queries, sampled from the \textsc{Beaver} and \textsc{Bird} datasets, and instructed to write a natural language description for each SQL query. The task was SQL-to-NL annotation only. A practical constraint in this setting is the \emph{cold-start} condition: at the beginning of the study, no prior annotations exist, and thus the system cannot retrieve similar examples via RAG. Consequently, the first annotation is performed without example-based guidance, while subsequent annotations incrementally benefit from the examples created earlier in the session and the knowledge injected in the feedback-loop. This mirrors realistic enterprise deployments, where organizations typically begin with no internal annotation archive, and manually building such an example set would itself require substantial effort. BenchPress therefore assumes a cold start by default, while in future enterprise use cases we intend to offer both options: starting from an empty knowledge base or initializing the system with examples from public text-to-SQL datasets.

\textbf{Independent variables} in the study were the annotation condition (BenchPress, Manual, Vanilla LLM) and user expertise (Advanced, Non-Advanced). The primary \textbf{dependent variables} included annotation time and annotation quality. Quality was assessed using observational measures (e.g., back-translation match, ROUGE similarity). This setup enables a structured comparison of how different interfaces and user backgrounds affect performance in enterprise SQL annotation tasks.}


\subsection{Evaluation}

We evaluated the performance of each annotation condition -- \textit{BenchPress}, \textit{Manual}, and \textit{Vanilla LLM} -- across three dimensions: annotation accuracy, annotation latency, and semantic fidelity using a backtranslation task.

\paragraph{Annotation Accuracy.}
Annotation accuracy was measured by manually inspecting each NL description for fidelity to the corresponding SQL query. We checked whether key SQL components—such as column selections, calculations (e.g., aggregations), and grouping or ordering operations—were clearly and distinguishably described. As shown in Table~\ref{fig:accuracy}, BenchPress consistently outperformed the Manual and LLM groups across both datasets. It produced more complete and structurally accurate descriptions, especially in complex enterprise queries from the \textsc{Beaver} dataset.

\begin{table}[ht]
Avg Accuracy\\
\begin{tabular}{lccc}
    & \textbf{BenchPress } & \textbf{Vanilla LLM } & \textbf{Manual} \\
\hline
\textsc{Beaver}         & 86.1\%          & 66.2\%        & 60.1\%          \\
\textsc{Bird}            & 100.0\%          & 100.0\%        & 87.8\%          \\
\hline
Overall                  & 93.0\%          & 83.1\%        & 73.9\%          \\
\end{tabular}
    \centering
    \caption{Annotation accuracy across BenchPress, Vanilla LLM, and Manual conditions on \textsc{Beaver} and \textsc{Bird}.}
    \label{fig:accuracy}
\end{table}

\paragraph{Annotation Latency.}
Latency was measured as the total annotation time per participant, averaged across a given dataset. Table~\ref{fig:latency} shows that BenchPress led to the fastest annotation times, while the Manual group required by far the most time. This supports the hypothesis that context-aware LLM suggestions accelerate annotation without sacrificing quality.

\begin{table}[ht]

Avg Latency\\
\begin{tabular}{lccc}
    & \textbf{BenchPress } & \textbf{Vanilla LLM } & \textbf{Manual} \\
\hline
\textsc{Beaver}         & 16.1 min          & 16.2 min        & 102.1 min          \\
\textsc{Bird}            & 12.0 min         & 15.8 min        & 82.8 min          \\
\hline
Total                  & 28.1 min          & 32.0 min        & 183.9min          \\
\end{tabular}
    \centering
    \caption{Average annotation latency (in minutes) per condition across all participants for each dataset.}
    \label{fig:latency}
\end{table}

\paragraph{Annotation Fidelity via Backtranslation.}
To evaluate the semantic accuracy of the NL annotations produced in BenchPress, we performed a backtranslation task. In this process, an LLM was asked to regenerate SQL queries solely from the natural language descriptions created during annotation. The resulting SQL outputs were then compared to the original queries using a 5-level rubric.

We chose this rubric to capture both hard execution failures and finer semantic mismatches, providing a practical scale for measuring fidelity. Our rating system distinguishes between structural errors (e.g., incorrect tables or joins), content-level inaccuracies (e.g., wrong columns or filters), and minor deviations (e.g., ordering or phrasing issues). The five levels are defined as follows:

\begin{itemize}
    \item \textbf{Level 1 – Invalid:} The generated SQL fails to execute (e.g., due to syntax errors, undefined references, or broken nesting).
    
    \item \textbf{Level 2 – Executable but Structurally Incorrect:} The SQL query runs but reflects major misunderstandings of the query’s structure. Examples include wrong tables, missing joins, or irrelevant subqueries.
    
    \item \textbf{Level 3 – Column-Level Errors:} The SQL is structurally correct but uses incorrect columns, filters, functions, or groupings. The high-level intent is preserved, but the query’s logic is incorrect.
    
    \item \textbf{Level 4 – Minor Issues:} The regenerated SQL is mostly faithful but contains small deviations such as incorrect sorting, missing nuance, or redundant clauses.
    
    \item \textbf{Level 5 – Fully Correct:} The SQL matches the original in both structure and semantics, including all tables, conditions, filters, and ordering.
\end{itemize}

We used a vanilla LLM (without fine-tuning, chain-of-thought prompting, or in-context examples) to ensure that the results reflect the inherent information content of the NL descriptions, not artifacts of prompt engineering. This makes the evaluation a stricter test of how well the natural language alone communicates the SQL logic.

Backtranslation in this form provides a valuable lens on annotation quality: it captures not only whether a human would find a description understandable, but also whether it preserves enough detail for faithful round-tripping into executable SQL.

\begin{figure}[ht]
    \centering
    \includegraphics[width=\linewidth]{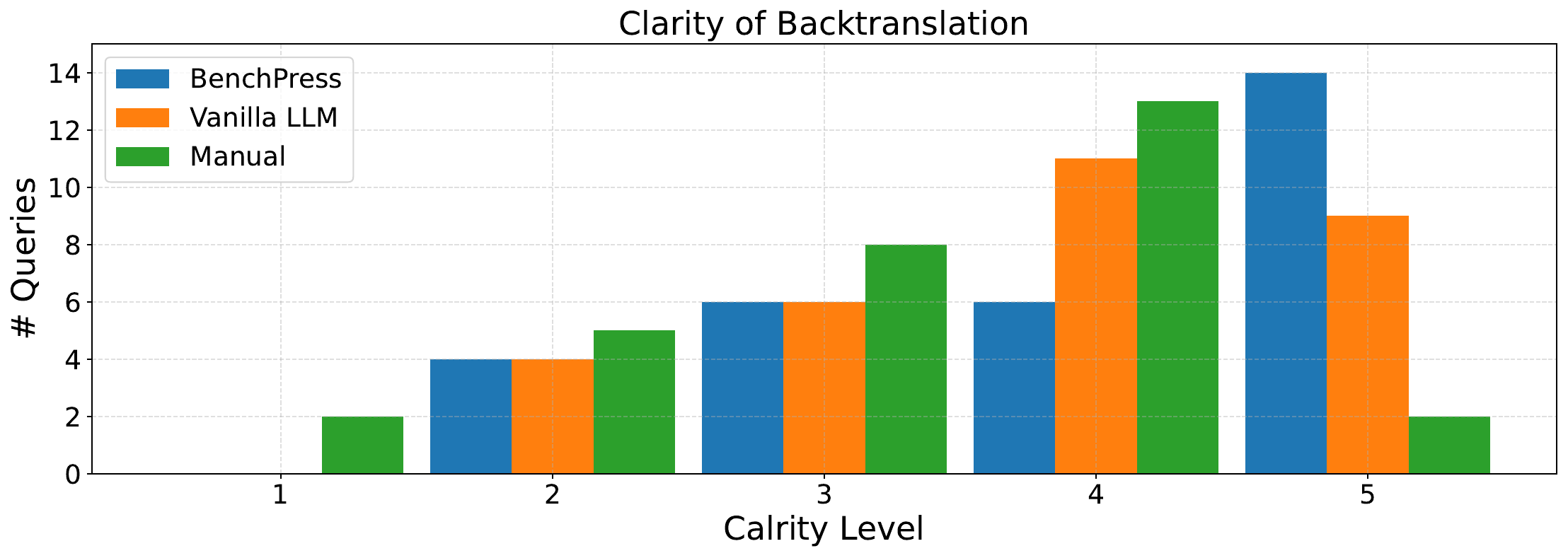}
    \caption{Backtranslation fidelity: proportion of SQL outputs at each clarity level across conditions.}
    \label{fig:backtranslation}
\end{figure}

Figure~\ref{fig:backtranslation} shows that BenchPress yielded the highest proportion of 5 outputs, indicating superior semantic clarity. Manual and LLM groups more often fell into Level 4, typically due to row-order inconsistencies, omitted nuances, or subtle misinterpretations of intent.

\subsection{Summary and Observations}

Our user study yielded several notable insights:

\paragraph{Tool Effectiveness.}
BenchPress consistently outperformed both the vanilla LLM and manual annotation approaches across all metrics—accuracy, efficiency, and semantic fidelity. These results demonstrate that integrating LLM-generated suggestions with lightweight user guidance significantly enhances the annotation process. The structured interface and contextual prompts provided by BenchPress not only improved output quality but also reduced cognitive load and task completion time.

\paragraph{Dataset Complexity.}
We observed a clear divergence in tool performance between public and enterprise datasets. While all three tools performed reasonably well on the \textsc{Bird} dataset, the \textsc{Beaver} dataset—with its higher schema complexity, ambiguous column names, and enterprise-specific terminology—exposed substantial differences. BenchPress maintained high annotation quality in this more challenging setting, whereas the manual and vanilla LLM conditions struggled to preserve SQL semantics and coverage.

\paragraph{Implications for Benchmarking.}
These findings highlight the limitations of existing tools and benchmarks when applied to real-world enterprise data. Public benchmarks do not adequately reflect the structural and linguistic complexity inherent in enterprise SQL logs. BenchPress addresses this gap by enabling scalable and accurate benchmark curation tailored to organizational data environments. By focusing on SQL-to-NL annotation with human-in-the-loop refinement, BenchPress provides a practical foundation for building robust, domain-specific text-to-SQL benchmarks and evaluating model performance in enterprise settings.

\new{\section{Enterprise SQL Logs: Contributions}

In this section, we highlight the contributions how \textsc{BenchPress} addresses challenges unique to enterprise SQL logs explicit by discussing (i) schema complexity, (ii) domain-specific terminology, and (iii) privacy and confidentiality constraints. As introduced in Section~\ref{sec:introduction}, these challenges are well-documented in enterprise workloads such as \textsc{Beaver}, and they directly inform the design of \textsc{BenchPress}.

\paragraph{Schema Complexity and Ambiguity}
As discussed in Section~\ref{sec:Enterprise SQL Logs: Challenges}, enterprise schemas often exhibit extensive duplication of names, values, and attribute types. This kind of structural redundancy and overlap is characteristic of large warehouse systems and rarely appears in public academic datasets. \textsc{BenchPress} mitigates this challenge by retrieving past SQL queries that use the same columns, thereby exposing annotators to the actual join patterns and filters associated with ambiguous attributes. For example, retrieved usage clarifies that \texttt{ACADEMIC\_TERMS\_ALL} contains both historical and current term records, whereas \texttt{ACADEMIC\_TERMS} focuses only on the current period. Similarly, filters such as \texttt{STREET\_TYPE = 'STREET'} in address tables indicate that only physical street addresses—not mailing addresses—should be considered. By surfacing these real usage patterns, \textsc{BenchPress} turns otherwise opaque schemas into interpretable annotation context and reduces confusion that would otherwise propagate into downstream benchmarks.

\paragraph{Domain-Specific Terminology and Abbreviations}
Enterprise SQL logs also contain specialized vocabulary from domains such as finance, HR, and student administration—terminology that is often absent from table documentation. \textsc{BenchPress} assists annotators by generating multiple candidate NL descriptions informed by retrieved query contexts. These candidates help annotators quickly infer the correct domain meaning while avoiding hallucinations, enabling effective annotation even when no glossary or documentation is available—a condition frequently noted by reviewers.

\paragraph{Privacy and Confidentiality Constraints}
Enterprise SQL logs also contain specialized vocabulary from domains such as finance, HR, and student administration—terminology that is often absent from table documentation. BenchPress assists annotators by generating multiple candidate NL descriptions informed by retrieved query contexts. These candidates help annotators quickly infer the correct domain meaning while avoiding hallucinations, enabling effective annotation even when no glossary or documentation is available—a condition frequently noted by reviewers. Moreover, through BenchPress’s iterative feedback loop, any domain-specific clarifications provided by annotators are fed back into the prompt for subsequent queries. This eliminates the need to repeatedly look up the same information in external documentation, as once a piece of knowledge has been captured, the system is able to reuse it automatically in future generations.

\paragraph{Connection to Enterprise Difficulty}
The \textsc{Beaver} benchmark demonstrates that state-of-the-art LLMs suffer a substantial execution accuracy drop (up to 70--90\%) when evaluated on enterprise SQL logs, despite strong performance on public datasets. Tables~\ref{tab:query-complexity} and \ref{tab:data-complexity} quantify these differences in query and data complexity.

The \textsc{Beaver} benchmark shows that state-of-the-art LLMs experience dramatic execution accuracy drops on enterprise workloads due to schema complexity, domain terminology, and data irregularity. Tables~\ref{tab:query-complexity} and \ref{tab:data-complexity} illustrate these differences. \textsc{BenchPress} is designed explicitly to help organizations create domain-specific text-to-SQL benchmarks that reflect these real-world complexities, thereby reducing deployment risks and improving model evaluation fidelity.

}


\section{Conclusions and Future Work}

BenchPress accelerates the creation of high-quality, domain-specific text-to-SQL benchmarks by combining LLM-based generation with human-in-the-loop validation. Our system uses retrieval-augmented prompting for intelligent SQL-to-text annotation and enables scalable evaluation of model performance on enterprise workloads. Through a controlled user study, we demonstrated that BenchPress improves annotation speed and fidelity, especially in typical enterprise settings with complex schemas and limited training data.

While the current system focuses on SQL-to-text annotation, a natural next step is to incorporate text-to-SQL generation for iterative validation. This would further increase the accuracy and speed of the benchmark curation process.

Another direction for future work is assessing the robustness of state-of-the-art models trained on public benchmarks such as \textsc{Spider}, \textsc{Bird}, and \textsc{Fiben}. Although many models achieve near-perfect performance on these datasets, it remains unclear whether they have overfit to canonical NL formulations. We plan to systematically rephrase the natural language queries in existing benchmarks—introducing more realistic, ambiguous, or underspecified variants—and re-evaluate model performance. This will reveal whether current benchmarks reflect genuine text-to-SQL generalization or merely reward surface-level pattern matching.

Creating custom benchmarks for domain-specific tasks with private data remains a significant bottleneck in the broader adoption of LLM applications. Data teams need better tools to rapidly create benchmarks that are representative of their unique data and use cases. BenchPress addresses this need by enabling adaptive, privacy-aware, and task-specific benchmarking of text-to-SQL systems within enterprise contexts.

\new{\section*{Acknowledgments}
We thank Michael Stonebraker, Peter Baile Chen, Bijan Arbab, Moh Haghighat and Layne Mills for their invaluable feedback throughout the development of BenchPress. Although not listed as co-authors due to potential conflicts with other CIDR submissions, their contributions were instrumental to this work. We also thank the participants of our user study for their time and effort, which helped us evaluate and refine the system. This research is supported by Intel as part of the MIT Data Systems and AI Lab (DSAIL).}

\bibliographystyle{abbrvnat}
\bibliography{bibliography-short.bib}
\end{document}